\documentclass[10pt,conference]{IEEEtran}
\IEEEoverridecommandlockouts
\usepackage{cite}
\usepackage{amsmath,amssymb,amsfonts}
\usepackage{algorithmic}
\usepackage{graphicx}
\usepackage{textcomp}
\usepackage{xcolor}
\def\BibTeX{{\rm B\kern-.05em{\sc i\kern-.025em b}\kern-.08em
    T\kern-.1667em\lower.7ex\hbox{E}\kern-.125emX}}
\begin{document}

\title{Causal Discovery for Cross-Sectional Data Based on Super-Structure and Divide-and-Conquer*\\
}
\author{\IEEEauthorblockN{1\textsuperscript{st} Wenyu Wang}
\IEEEauthorblockA{\textit{School of Computer Science} \\
\textit{University of South China}\\
Hengyang, China \\
20232008110502@stu.usc.edu.cn}
\and
\IEEEauthorblockN{2\textsuperscript{nd} Yaping Wan}
\IEEEauthorblockA{\textit{School of Computer Science} \\
\textit{University of South China}\\
Hengyang, China \\
ypwan@usc.edu.cn}
}

\maketitle

\begin{abstract}

This paper tackles a critical bottleneck in Super-Structure-based divide-and-conquer causal discovery: the high computational cost of constructing accurate Super-Structures—particularly when conditional independence (CI) tests are expensive and domain knowledge is unavailable. We propose a novel, lightweight framework that relaxes the strict requirements on Super-Structure construction while preserving the algorithmic benefits of divide-and-conquer. By integrating weakly constrained Super-Structures with efficient graph partitioning and merging strategies, our approach substantially lowers CI test overhead without sacrificing accuracy.
We instantiate the framework in a concrete causal discovery algorithm and rigorously evaluate its components on synthetic data. Comprehensive experiments on Gaussian Bayesian networks, including magic-NIAB, ECOLI70, and magic-IRRI, demonstrate that our method matches or closely approximates the structural accuracy of PC and FCI while drastically reducing the number of CI tests. Further validation on the real-world China Health and Retirement Longitudinal Study (CHARLS) dataset confirms its practical applicability. Our results establish that accurate, scalable causal discovery is achievable even under minimal assumptions about the initial Super-Structure, opening new avenues for applying divide-and-conquer methods to large-scale, knowledge-scarce domains such as biomedical and social science research.
\end{abstract}


\section{Introduction}

Causal discovery aims to recover a directed acyclic graph (DAG) from observational data, enabling mechanistic interpretation and downstream reasoning. In practice, however, learning causal structure at scale remains challenging. Mainstream paradigms include constraint-based methods~\cite{spirtes_notitle_2000}, score-based search~\cite{chickering_optimal_2005}, and hybrid approaches~\cite{tsamardinos_max-min_2006}. When the number of variables grows, these methods often become dominated by conditional independence (CI) testing or combinatorial search over candidate graphs, leading to substantial computational overhead and degraded reliability~\cite{bergsma_testing_2004}.

To improve scalability, divide-and-conquer strategies have become increasingly attractive: they first partition variables into smaller subsets, learn local subgraphs, and then merge them into a global structure. A representative direction is to introduce a Super-Structure to constrain candidate edges and guide graph partitioning~\cite{shah_causal_2025}. Yet existing Super-Structure-based pipelines typically require the Super-Structure to contain (or closely approximate) the true skeleton, which implicitly demands high recall. Constructing such a high-recall scaffold without domain knowledge is expensive, since it still requires many CI tests or heavy pre-processing, and the construction cost can erase the downstream gains.

This paper proposes a lightweight alternative by weakening the constraint on Super-Structures. Instead of requiring the true skeleton to be fully contained in the Super-Structure, we only require the Super-Structure to be a subgraph of the true skeleton. This shifts the design goal from high recall to high precision: the Super-Structure may miss edges, but the edges it includes are reliable and therefore safe to use for partitioning. Conceptually, this is analogous to a principle that has proven effective in other large-scale learning systems: injecting a small amount of reliable structure can dramatically reduce search complexity while keeping overall quality. For example, in controllable diffusion and conditional generation, explicitly enforcing reliable constraints (e.g., consistent quantity and layout) can stabilize optimization and reduce ambiguity in complex editing tasks~\cite{shen2025imagharmony}. Similarly, attribute-wise or modular conditioning can improve controllability and efficiency by restricting the hypothesis space to the most informative factors~\cite{shen2025imaggarment, shen2025imagdressing}. Progressive conditioning and unified pose-guided formulations further illustrate how structured intermediate representations can improve robustness under challenging variations~\cite{shen2024advancing, shen2024imagpose}. Rich-context conditioning has also been shown to reduce long-range inconsistency in sequential generation by providing a more reliable contextual scaffold~\cite{shen2025boosting}, while motion priors can stabilize long-horizon temporal coherence by constraining dynamics~\cite{shenlong}. Inspired by these successes, we apply the same structural viewpoint to causal discovery: a weak but reliable graph scaffold can enable efficient decomposition without requiring an expensive, high-recall pre-estimation step.

We instantiate the proposed framework with a concrete algorithm and validate it on synthetic benchmarks and real-world datasets, including large-scale biomedical and social science data. Experiments demonstrate that our method substantially reduces the number of CI tests while maintaining competitive structural accuracy, showing that weakly constrained Super-Structures can deliver meaningful computational savings with minimal loss in fidelity.

Our contributions are:
\begin{itemize}
\item We introduce a causal partitioning framework that integrates weakly constrained Super-Structures with divide-and-conquer causal discovery, enabling efficient dimensionality reduction with a low-cost scaffold.
\item We present an algorithmic realization that operationalizes this framework and provides practical efficiency improvements in high-dimensional settings.
\item We provide empirical evidence on both synthetic and real datasets demonstrating large reductions in CI tests with minimal accuracy degradation.
\end{itemize}

\section{Related Work}

Virtually all causal discovery algorithms operate by searching within a structured hypothesis space of directed acyclic graphs (DAGs). Divide-and-conquer approaches accelerate this search by partitioning the variable set into subsets, performing local causal discovery on each, and then merging the results to reconstruct the full causal graph.
Causal discovery methods based on the divide-and-conquer approach can be categorized into three types:
\subsubsection{Recursive CI-based methods}
Cai et al.~\cite{cai_sada_2013} introduced SADA, a recursive algorithm that, at each step, uses either multiple low-order or a single high-order CI test to infer a decomposition structure and derive variable subsets. However, the resulting subset sizes are uncontrolled and must be enumerated to meet target constraints. Zhang~\cite{zhang_learning_2022} proposed CAPA, another recursive method that partitions variables into two subsets per recursion using only low-order CI tests, combined with a regression-based CI test tailored for linear non-Gaussian additive noise models to preserve d-separation~\cite{pearl_notitle_2000}.
\subsubsection{Clustering-based approaches}
which partitions variables—treated as nodes—into clusters via hierarchical clustering, validated on continuous data. Huang \& Zhou extended this with pHGS~\cite{huang_partitioned_2022}, adapting PEF for discrete data. Chen et al.~\cite{chen_cdsc_2024} proposed CDSC, leveraging spectral clustering for dimensionality reduction in high-dimensional settings, while Wei et al.~\cite{wei_cdkm_2025} developed CDKM, a K-means–based method using minimum correlation distance for variable clustering.
\subsubsection{Super-Structure–based methods}
By leveraging a Super-Structure, the causal partitioning problem can be naturally cast as graph partitioning. Zhang~\cite{zhang_towards_2024} proposed CPA, which reformulates causal partitioning as edge-cutting on the adjacency graph, stably splitting variables into two subsets. Building on the same foundation, Shah et al.~\cite{shah_causal_2025} introduced the Expansive Causal Partition framework, which unifies multiple graph partitioning strategies to achieve flexible and scalable causal decomposition.

Perrier~\cite{perrier_finding_2008} introduced the Super-Structure—a scaffold for causal search constructed via low-order CI tests and/or domain knowledge—to reduce computational complexity. While Super-Structures can also be derived from CI testing or constraint-based methods, these alternatives incur substantial overhead in high-dimensional settings. To address this, we propose a weakly constrained Super-Structure, which relaxes construction requirements while retaining compatibility with the Super-Structure framework. Our approach supports diverse construction strategies—including domain knowledge, correlation measures, and score-based causal discovery—with details provided in \textbf{Section III}.


\section{Method}
\begin{figure}[htbp]
    \centering
    \includegraphics[width=0.6\linewidth, height=0.48\linewidth]{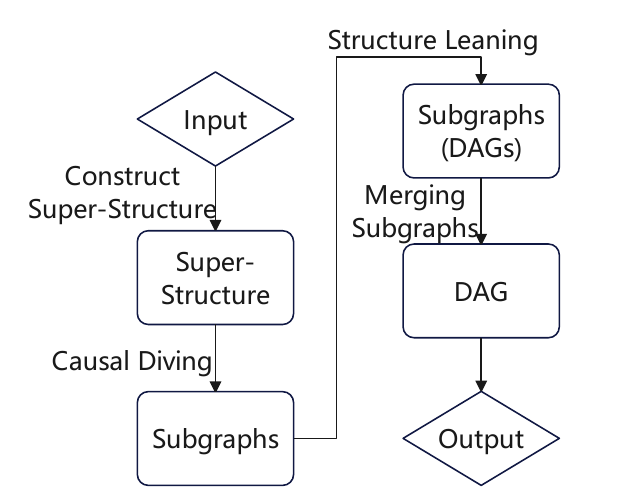}
    \caption{Pipeline overview figure.}
    \label{fig:pipe}
\end{figure}

Following Perrier~\cite{perrier_finding_2008}, a Super-Structure that contains the true skeleton is deemed sound; otherwise, it is incomplete. In this work, we adopt the converse convention for weakly constrained Super-Structures: those contained within the true skeleton are labeled sound, and others incomplete. Prior methods require highly sound Super-Structures to preserve d-separation across subgraphs and minimize redundant CI tests—yet their construction heavily relies on domain knowledge. In high-dimensional, mixed-distribution settings without such priors, achieving soundness becomes computationally prohibitive.
To overcome this, we embrace a lightweight strategy that deliberately uses incomplete weakly constrained Super-Structures, sidestepping the burden of constructing fully sound ones. This approach reframes the problem as optimizing the algorithm’s initial search point: as long as the weakly constrained Super-Structure is closer to the true graph than the naive starting point (e.g., a complete graph), it effectively reduces computational overhead. By leveraging such Super-Structures, we apply low-cost graph partitioning to decompose the problem into smaller subgraphs—dramatically lowering dimensionality, avoiding expensive high-order CI tests, and shrinking the search space. Building on this insight, we propose a divide-and-conquer causal discovery framework; an overview of the pipeline is shown in Fig.~\ref{fig:pipe}.

In the upcoming content of this section, we will demonstrate how to build a causal discovery algorithm based on the proposed framework.

\subsection{Super-Structure constructing module}

This module takes a dataset as input and outputs a weakly constrained Super-Structure. To construct an extreme contrast to Perrier’s “sound” Super-Structure~\cite{perrier_finding_2008}, we employ the Chow–Liu algorithm~\cite{chow_approximating_1968}, which builds a maximum spanning tree (MST) over a dependency matrix to yield a sparse, tree-structured approximation of the underlying graph. The procedure consists of: (i) constructing a pairwise dependency matrix, and (ii) computing its MST.

We replace the original Pearson correlation with Copula entropy~\cite{jian_mutual_2011}—a model-free, non-parametric dependence measure—to better capture complex, non-Gaussian relationships. Formally, for random variables $X$ with marginal distributions and copula density $c(u)$ , Copula entropy quantifies dependence without distributional assumptions, enhancing robustness across diverse data types. The copula entropy of $X$ is
\begin{equation}
H_{c}\left(X\right)=-\intop\limits_{u}{c\left(u\right)\log{c\left(u\right)}\mathrm{d}u}
\end{equation}

\subsection{Causal dividing module}

\begin{figure}[htbp]
    \centering
    \includegraphics[width=0.7\linewidth, height=0.35\linewidth]{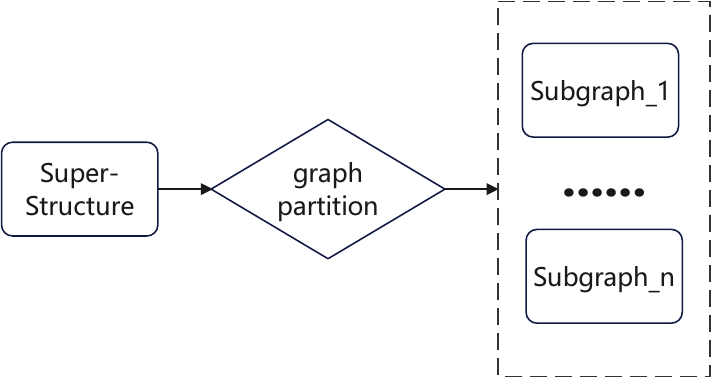}
    \caption{Causal dividing module.}
    \label{fig:graph-partitioning}
\end{figure}

This module takes a Super-Structure as input and outputs the segmented subgraphs, with the specific process shown in Fig.~\ref{fig:graph-partitioning}.
Here we use Girvan's method~\cite{girvan_community_2002} to implement graph partitioning and use causal extension method~\cite{shah_causal_2025} to process the subgraphs after partitioning.
This module partitions nodes into subgraphs via the Super-Structure, reducing dimensionality and computational cost. However, the weakly constrained Super-Structure may omit the true skeleton, violating d-separation across subgraphs and necessitating additional CI tests during merging—partially eroding the initial computational gains.

\subsection{Subgraph Learning module}

This module takes the subgraphs from the previous stage as input and outputs a skeleton graph. It employs a basic two-phase search strategy initialized from the subgraph as the starting point: (1) a \textbf{forward phase}, which iteratively tests all untested non-adjacent node pairs and adds edges for those exhibiting dependence; and (2) a \textbf{backward phase}, which re-evaluates all adjacent pairs and removes edges deemed conditionally independent.

\subsection{Subgraph merging module}

The input of this module is the learning results of the previous module, and the merged DAG is the output. In this article, this module first completes the correction by traversing all pairs of untested nodes for CI using the same strategy as the Subgraph Learning module, and then implements the merging using Shah's method~\cite{shah_causal_2025}.The specific process is shown as follows:

\begin{figure}[htbp]
    \centering
    \includegraphics[width=0.7\linewidth, height=0.35\linewidth]{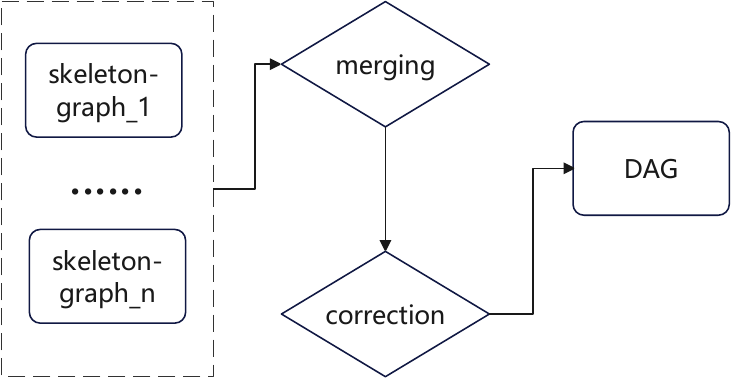}
    \caption{Subgraph merging module.}
    \label{fig:merging}
\end{figure}

\section{Experiments}

\subsection{Evaluation metrics}
Causal discovery aims to recover causal relationships, and its evaluation metrics typically measure the accuracy of inferred causal structures. We adopt the following metrics: Structural Hamming Distance (SHD), F1-score, Precision, Recall, Accuracy, and CI test count—where CI test denotes the number of unique CI tests executed during skeleton learning. Under the three conditions outlined earlier, repeated CI tests are cached, rendering their time cost negligible; thus, skeleton learning constitutes the primary phase where constraint-based algorithms perform CI tests.
Given that this study focuses on the skeleton learning phase, our experiments also concentrate on this aspect. The evaluation metrics above are applied to skeleton-graphs rather than DAGs.

\subsection{Simulation Experiment}
\subsubsection{CD Module Ablation Experiment}
This experiment evaluates the impact of the Causal Dividing (CD) module on algorithm performance. We compare a partial variant (without subgraph partitioning) against the full algorithm on synthetic data, and further assess performance when initializing the Super-Structure with maximum spanning trees derived from weight matrices built using different dependence measures.Synthetic datasets are generated following Strobl's method~\cite{strobl_fast_2018}: for a given node count $p$ , we construct a random DAG with expected indegree $E(n)=0.3795p$. Specifically, a lower-triangular adjacency matrix is sampled from $Bernoulli(0.075p/(p-1))$ ,then randomly permuted to break topological order. Edge weights are assigned i.i.d. from $Uniform[0.5,0.9]$ to form the weighted adjacency matrix, which defines linear Gaussian SEM coefficients.To isolate the CD module’s effect, we generate datasets with $p\in\{20,22,…,40\}$ , each with 5,000 samples and Gaussian noise. Fisher’s Z-test is used for CI testing. For each p , five graph instances are created; both algorithm variants are run four times per instance, and metrics are averaged. Results are summarized below:

\begin{figure}[htbp]
    \centering
    \includegraphics[width=0.8\linewidth, height=0.4\linewidth]{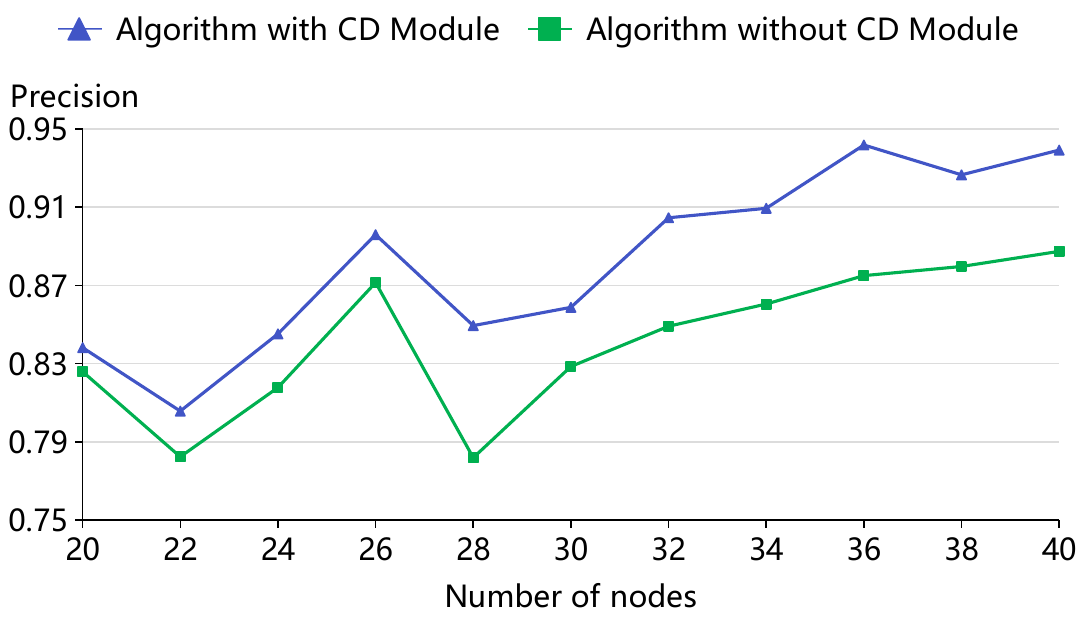}
    \caption{Precision of ablation experiment.}
    \label{fig:1}
\end{figure}
\begin{figure}[htbp]
    \centering
    \includegraphics[width=0.8\linewidth, height=0.4\linewidth]{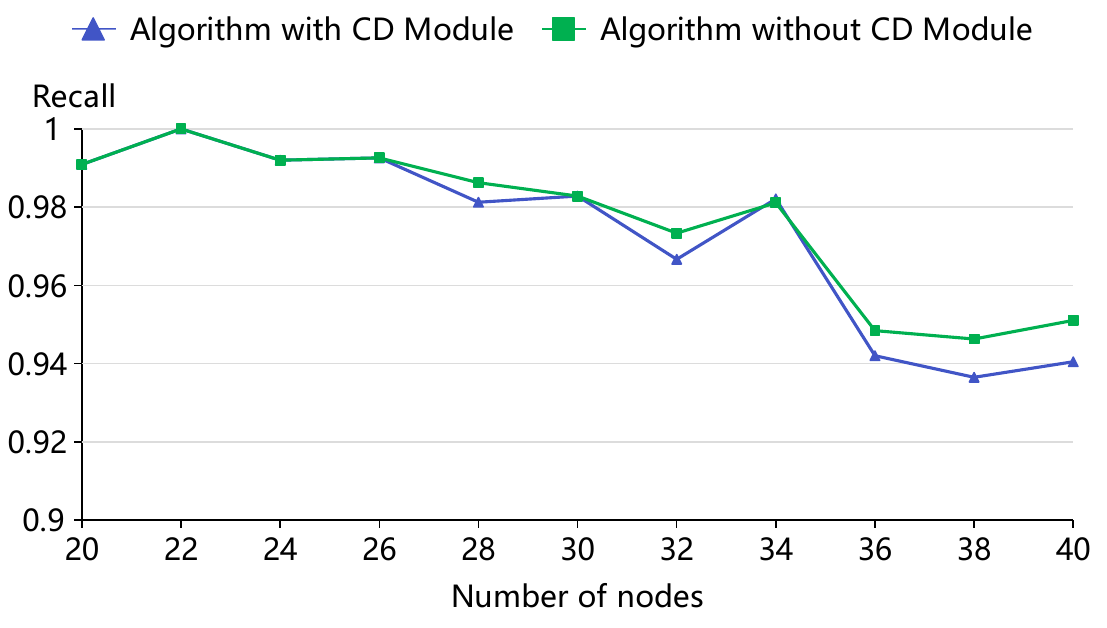}
    \caption{Recall of ablation experiment.}
    \label{fig:2}
\end{figure}

\begin{figure}[htbp]
    \centering
    \includegraphics[width=0.8\linewidth, height=0.4\linewidth]{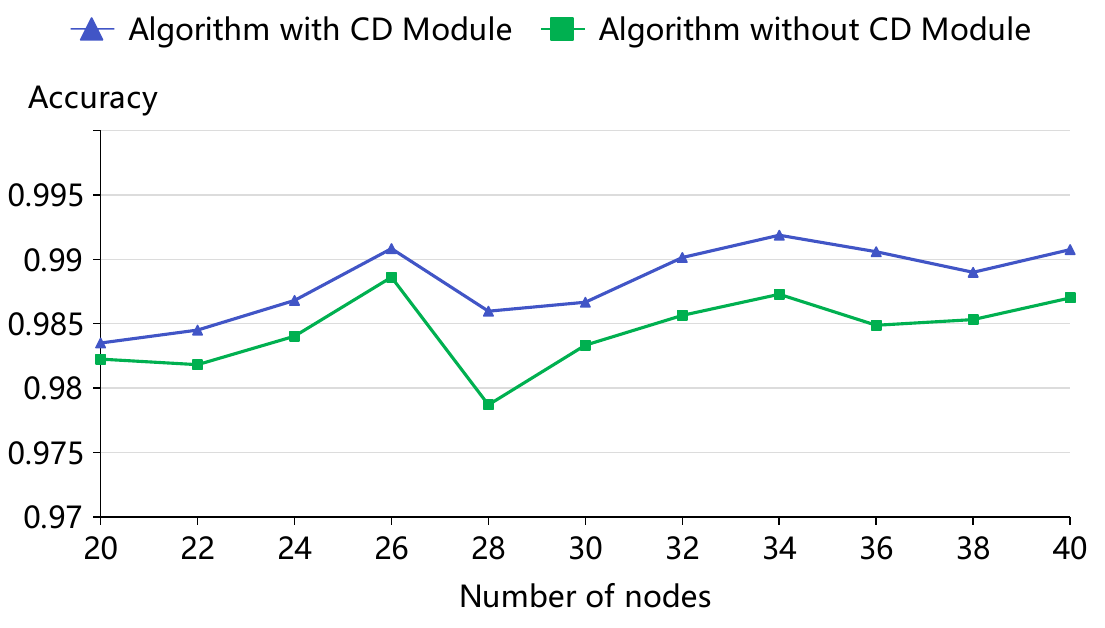}
    \caption{Accuracy of ablation experiment.}
    \label{fig:3}
\end{figure}
\begin{figure}[htbp]
    \centering
    \includegraphics[width=0.8\linewidth, height=0.4\linewidth]{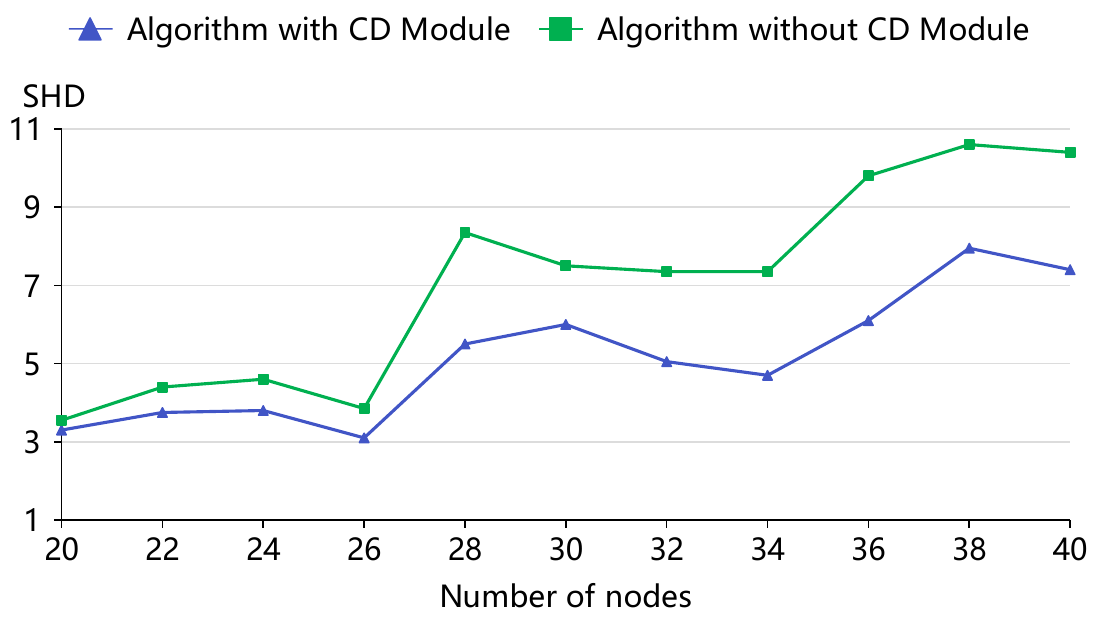}
    \caption{Structural Hamming Distance of ablation experiment.}
    \label{fig:4}
\end{figure}
\begin{figure}[htbp]
    \centering
    \includegraphics[width=0.8\linewidth, height=0.4\linewidth]{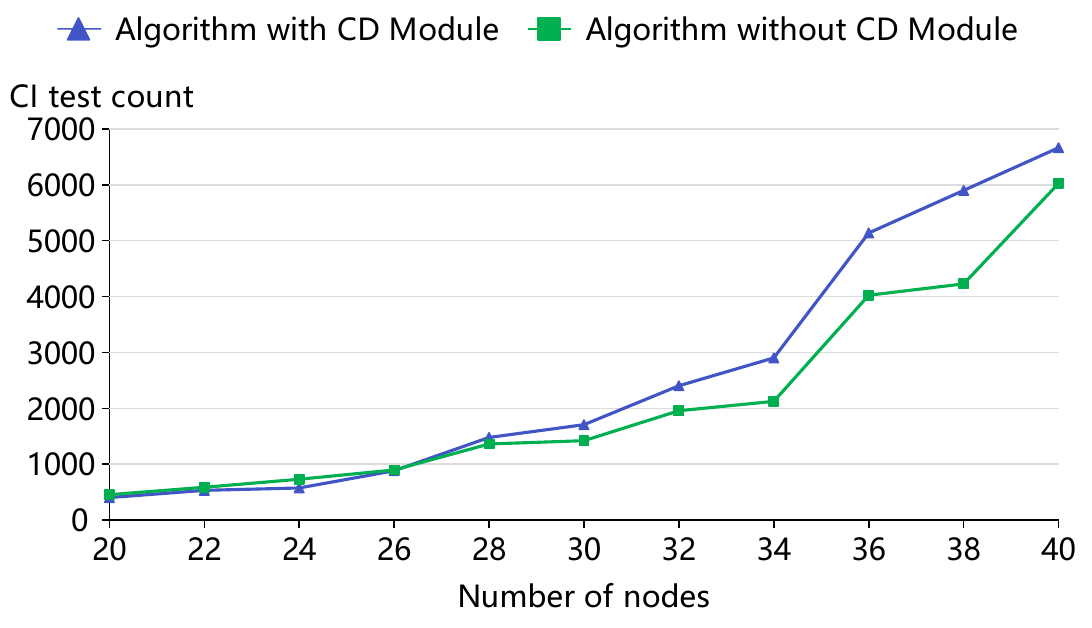}
    \caption{CI test count of ablation experiment.}
    \label{fig:5}
\end{figure}
\begin{figure}[htbp]
    \centering
    \includegraphics[width=0.8\linewidth, height=0.4\linewidth]{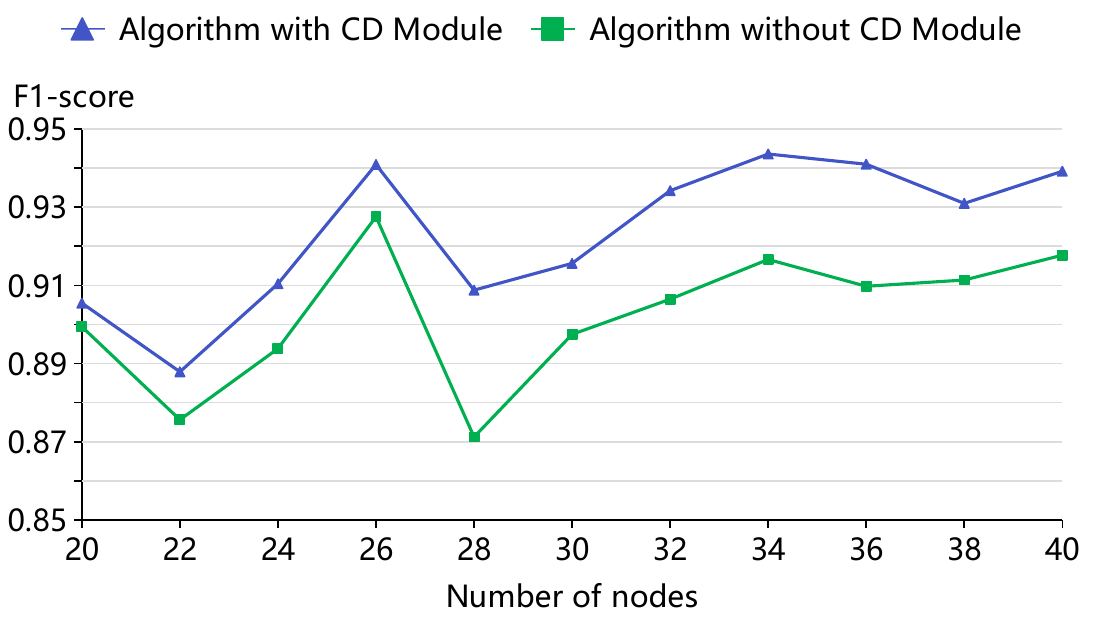}
    \caption{F1-score of ablation experiment.}
    \label{fig:6}
\end{figure}
Fig.~\ref{fig:1}$\sim$Fig.~\ref{fig:6}: The causal dividing (CD) module ablation experiment results. The x-axis shows the number of nodes, and the y-axis corresponds to the indicators shown in the legend. The green legend indicates the results without using the CD module, and the blue legend indicates the results using the CD module.

Experimental results show that the algorithm with the CD module outperforms the ablated version on most metrics—exhibiting only marginally higher CI test counts and slightly lower recall—demonstrating improved accuracy with minimal runtime overhead. This gain becomes more pronounced as the graph size increases, confirming that reduction in causal subset dimensionality through graph partitioning under a weakly constrained superstructure framework effectively enhances accuracy. The increased CI test count is primarily attributed to d-separation violations, which necessitate additional CI corrections. Given the current superstructure construction’s limited fidelity, refining it to better approximate the true skeleton is expected to substantially reduce this computational overhead.

\subsubsection{Correlation Metrics Comparison Experiment}
This experiment evaluates the impact of different correlation metrics on algorithm performance when constructing weakly constrained Super-Structures. We generated synthetic datasets with 24 nodes and 5,000 samples under four noise distributions—Gaussian, exponential, gamma, and uniform—producing two datasets per distribution. Using Petersen’s non-parametric CI test~\cite{petersen_testing_2021}, we compared four correlation measures: Copula entropy, mutual information, Pearson, and Spearman coefficients. Each dataset was run twice, and results were averaged across runs. The findings are summarized below:

\begin{figure}[htbp]
    \centering
    \includegraphics[width=0.8\linewidth, height=0.4\linewidth]{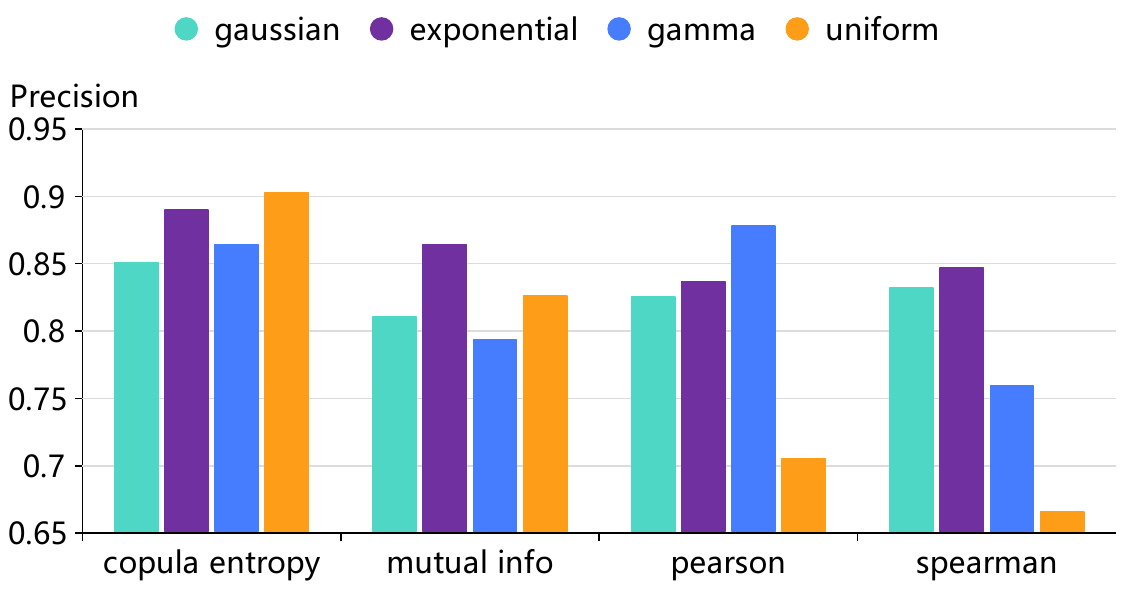}
    \caption{Precision of correlation metrics comparison experiment.}
    \label{fig:8}
\end{figure}
\begin{figure}[htbp]
    \centering
    \includegraphics[width=0.8\linewidth, height=0.4\linewidth]{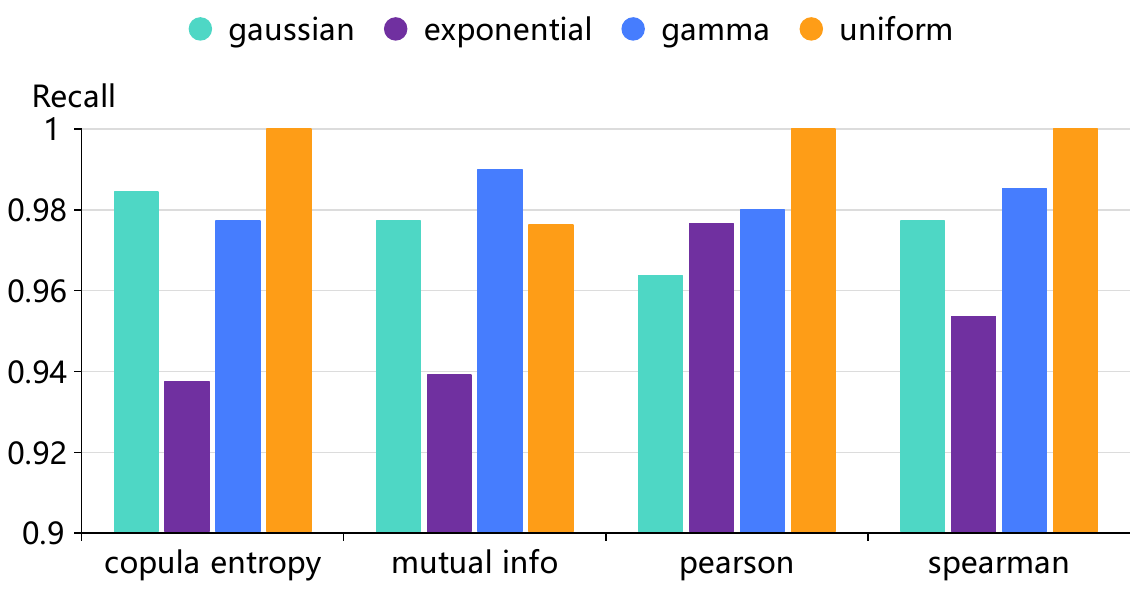}
    \caption{Recall of correlation metrics comparison experiment.}
    \label{fig:10}
\end{figure}

\begin{figure}[htbp]
    \centering
    \includegraphics[width=0.8\linewidth, height=0.4\linewidth]{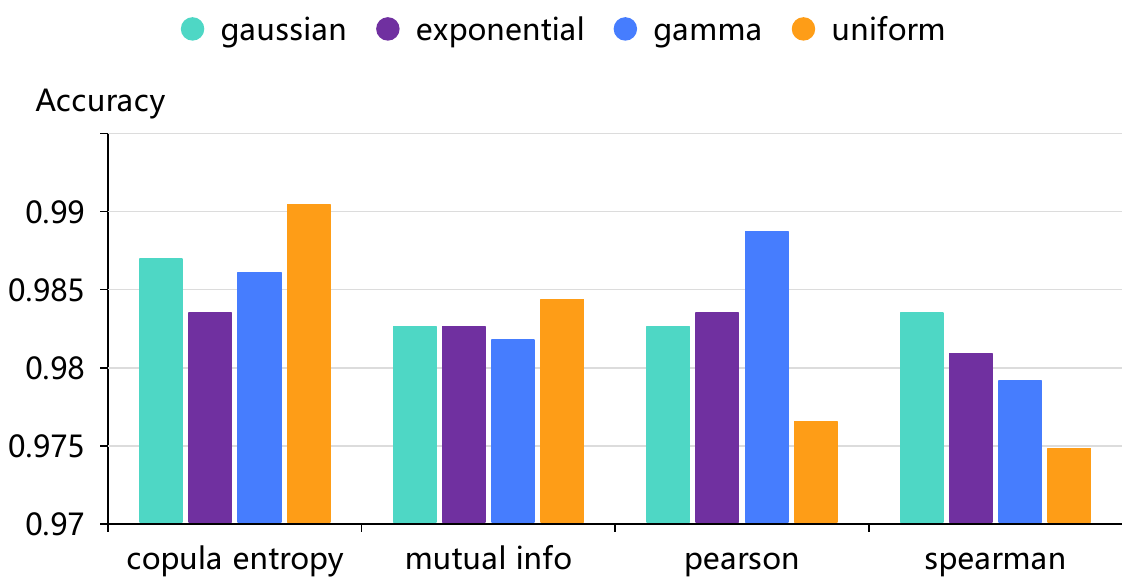}
    \caption{Accuracy of correlation metrics comparison experiment.}
    \label{fig:9}
\end{figure}
\begin{figure}[htbp]
    \centering
    \includegraphics[width=0.8\linewidth, height=0.4\linewidth]{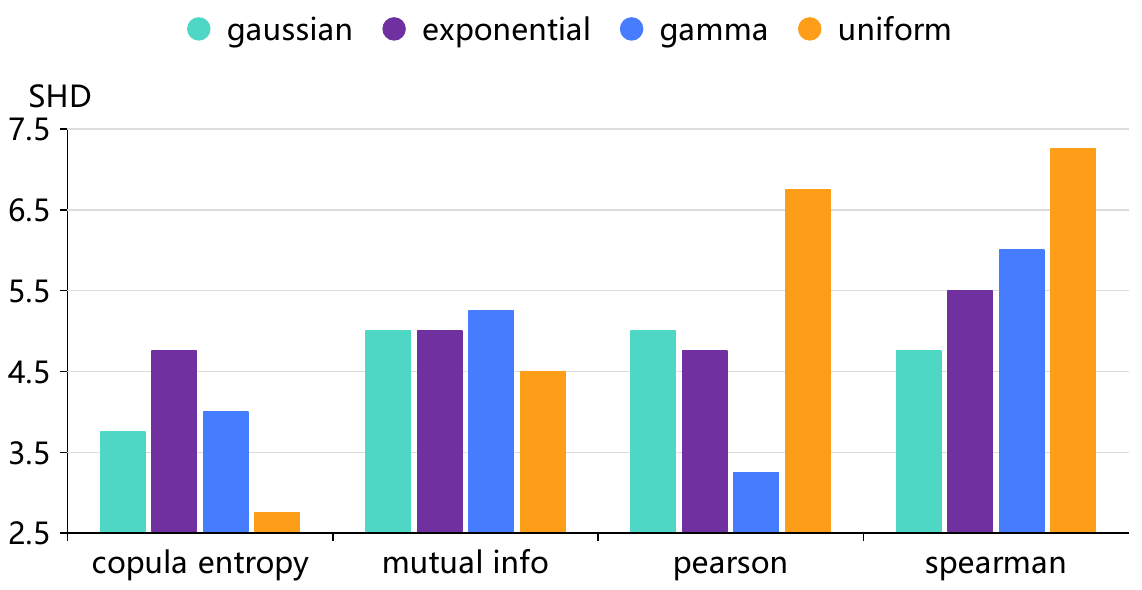}
    \caption{Structural Hamming Distance of correlation metrics comparison experiment.}
    \label{fig:12}
\end{figure}
\begin{figure}[htbp]
    \centering
    \includegraphics[width=0.8\linewidth, height=0.4\linewidth]{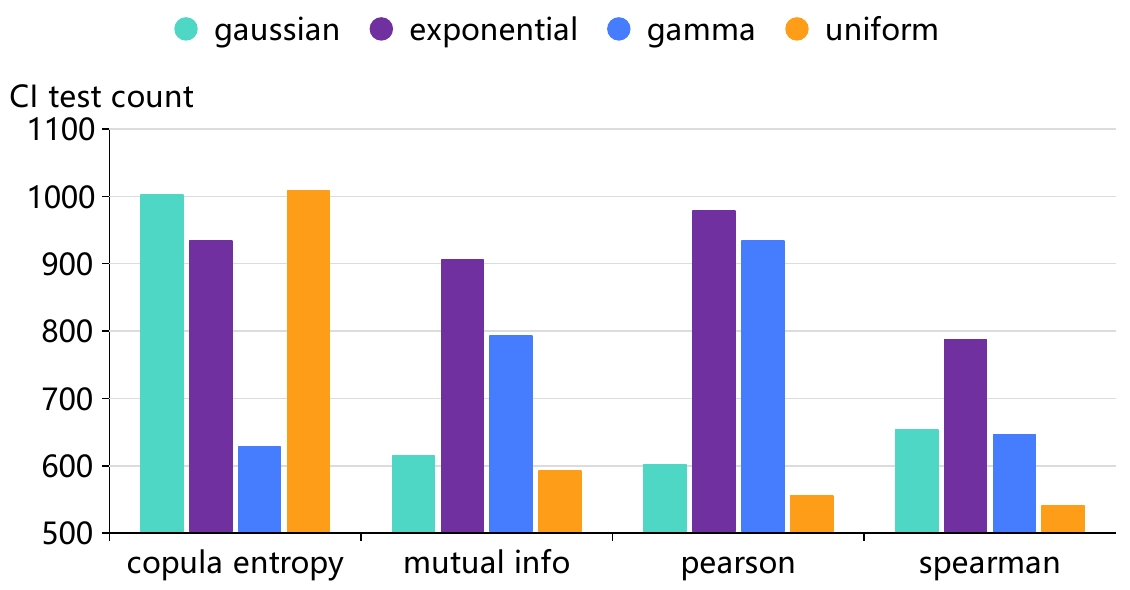}
    \caption{CI test count of correlation metrics comparison experiment.}
    \label{fig:11}
\end{figure}
\begin{figure}[htbp]
\centerline{\includegraphics[width=0.8\linewidth, height=0.4\linewidth]{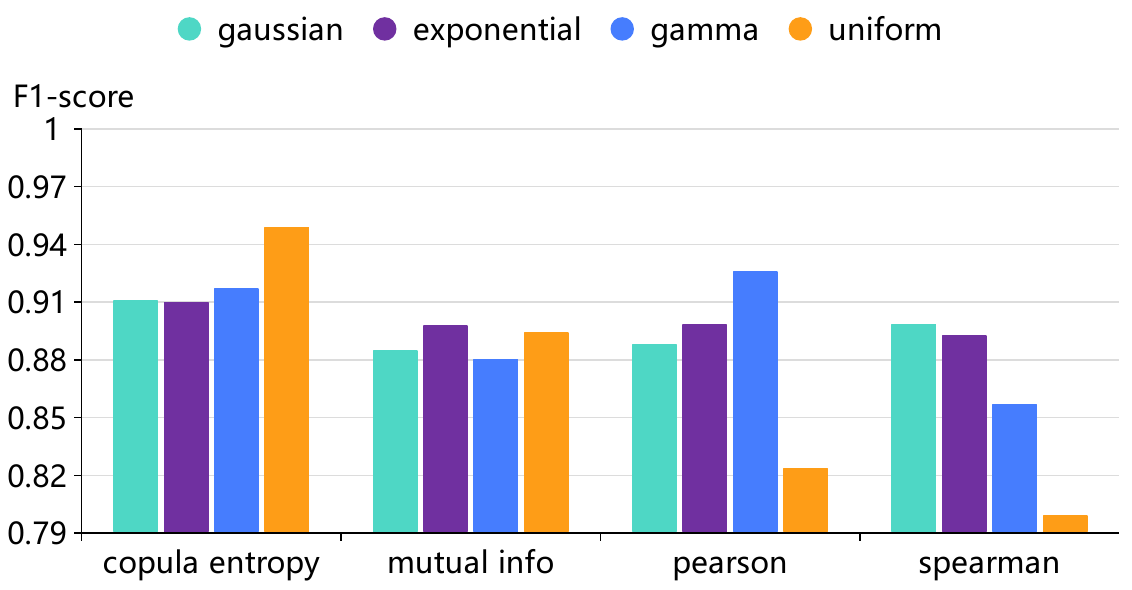}}
\caption{F1-score of correlation metrics comparison experiment.}
\label{fig:7}
\end{figure}
Fig.~\ref{fig:8}$\sim$Fig.~\ref{fig:7}: Experiment on effectiveness correlation metrics. The x-axis correspond to the metrics indicated in the legends, and the y-axes represents the correlation metrics from left to right: Copula Entropy, Mutual Information, Pearson coefficient, and Spearman coefficient. The green, purple, blue, and orange legends correspond to Gaussian, exponential, gamma, and uniform distributions, respectively.

The comparison experiment shows that, among correlation measures,\textbf{Copula entropy} yields the most robust Super-Structure construction—enabling our algorithm to better handle datasets with heterogeneous noise distributions. It outperforms alternatives on all metrics except recall and total CI tests, establishing it as the optimal choice for this task. Further experiments confirm that Copula entropy achieves the highest cross-distribution robustness and effectively captures dominant dependencies in mixed-distribution data.

\subsubsection{Algorithm Performance Comparison Experiment}
We use Gaussian Bayesian networks MAGIC-NIAB~\cite{scutari_multiple_2014}, ECOLI70~\cite{schafer_shrinkage_2005}, and MAGIC-IRRI~\cite{scutari_multiple_2014} to sample and generate simulated data to test the performance of our algorithm, with parameters as shown below.

\begin{table}[htbp]
\caption{Properties of Bayesian networks}
\begin{center}
\begin{tabular}{|c|c|c|c|}
\hline
\textbf{Bayesian networks} & MAGIC-NIAB &	ECOLI70 &	MAGIC-IRRI \\
\cline{1-4}
\textbf{Number of nodes} & 44 &	46 &	64 \\
\cline{1-4}
\textbf{Number of arcs} & 66 &	70 &	102 \\
\cline{1-4}
\textbf{Average Markov blanket} & 9.88	& 3.65	& 9.97 \\
\cline{1-4}
\textbf{Average degree} & 3	& 3.04	& 3.19 \\
\cline{1-4}
\textbf{Maximum in-degree} & 9	& 4	& 11 \\
\hline
\end{tabular}
\label{tab11}
\end{center}
\end{table}

Data were generated using the bnlearn R package by sampling Gaussian Bayesian networks from .rdn files. Each network produced 10 test sets of 5,000 samples each; every set was evaluated four times, and average performance metrics were reported. CI tests used Fisher’s Z without prior knowledge. Baseline methods were FCI and PC; our method is denoted Ag. Results are presented below:
\begin{table}[htbp]
\caption{Experimental results from MAGIC-NIAB}
\begin{center}
\begin{tabular}{|c|c|c|c|}
\hline
\textbf{Algorithm} & \textbf{Ag}	&	\textbf{FCI}	&	\textbf{PC}	\\
\cline{1-4}
\textbf{Accuracy} & 0.9840 &	0.9846 &	0.9841 \\
\cline{1-4}
\textbf{Precision} & 0.8108	& 0.8176	& 0.8124 \\
\cline{1-4}
\textbf{Recall} & 1.0000	& 1.0000	& 1.0000 \\
\cline{1-4}
\textbf{F1-score} & 0.8952	& 0.8992	& 0.8961 \\
\cline{1-4}
\textbf{SHD} & 15.5	& 14.9	& 15.4 \\
\cline{1-4}
\textbf{CI test count} & 16366.4	& 94287.3	& 18893.8 \\
\hline
\end{tabular}
\label{tab12}
\end{center}
\end{table}
\begin{table}[htbp]
\caption{Experimental results from ECOLI70}
\begin{center}
\begin{tabular}{|c|c|c|c|}
\hline
\textbf{Algorithm} & \textbf{Ag}	&	\textbf{FCI}	&	\textbf{PC}	\\
\cline{1-4}
\textbf{Accuracy} & 0.9856	& 0.9863 & 0.9861 \\
\cline{1-4}
\textbf{Precision} & 0.8672	& 0.8841 & 0.8803 \\
\cline{1-4}
\textbf{Recall} & 0.9396 & 0.9314 & 0.9350 \\
\cline{1-4}
\textbf{F1-score} & 0.8979 & 0.9018	& 0.9016 \\
\cline{1-4}
\textbf{SHD} & 14.5	& 13.9 & 14.0 \\
\cline{1-4}
\textbf{CI test count} & 11633.0	& 77255.9	& 27422.5 \\
\hline
\end{tabular}
\label{tab13}
\end{center}
\end{table}
\begin{table}[htbp]
\caption{Experimental results from MAGIC-IRRI}
\begin{center}
\begin{tabular}{|c|c|c|c|}
\hline
\textbf{Algorithm} & \textbf{Ag}	&	\textbf{FCI}	&	\textbf{PC}	\\
\cline{1-4}
\textbf{Accuracy} & 0.9856 & 0.9860 & 0.9858 \\
\cline{1-4}
\textbf{Precision} & 0.8429 & 0.8544 & 0.8501 \\
\cline{1-4}
\textbf{Recall} & 0.9475 & 0.9406 & 0.9429 \\
\cline{1-4}
\textbf{F1-score} & 0.8888 & 0.8908 & 0.8897 \\
\cline{1-4}
\textbf{SHD} & 19.5 & 19.1 & 19.4 \\
\cline{1-4}
\textbf{CI test count} & 17303.4 & 199399.2 & 31921.0 \\
\hline
\end{tabular}
\label{tab14}
\end{center}
\end{table}

Experimental results show that, except on the largest network (MAGIC-IRRI), our method matches baseline accuracy—exhibiting only slightly lower precision and F1, and modestly higher SHD and recall. All methods perform better on the sparser ECOLI70 than on MAGIC-NIAB. Crucially, our approach requires significantly fewer non-redundant CI tests across all networks, fulfilling its core design objective.These results indicate that even with a less accurate Super-Structure, overall causal learning performance remains robust. As a divide-and-conquer algorithm prioritizing CI test reduction over peak accuracy, our method successfully validates its central premise: relaxing strict Super-Structure constraints preserves correctness while enabling broader applicability.

\subsubsection{Real-world data}
We applied our method to cognitive function data from middle-aged and older adults in the China Health and Retirement Longitudinal Study (CHARLS)~\cite{zhao_cohort_2014}. Inclusion criteria required participants to: (i) complete the global cognitive assessment and depression questionnaire in Wave 3; and (ii) have completed the cognitive test in Wave 2 and participated in the Wave 3 physical exam and second blood draw. Exclusion criteria were self-reported dementia, Parkinson’s disease. Variable definitions are provided in the table below.
\begin{table}[htbp]
\caption{Variable names and definitions}
\begin{center}
\begin{tabular}{|c|c|}
\hline
\textbf{Variable} & \textbf{Definition}	\\
\cline{1-2}
X1 & Depression \\
X2 & Cognitive Function  \\
X3 & White Blood Cell  \\
X4 & Hemoglobin  \\
X5 & Hematocrit  \\
X6 & Mean Corpuscular Volume  \\
X7 & Platelets  \\
X8 & Triglycerides  \\
X9 & Creatinine  \\
X10 & Blood Urea Nitrogen  \\
X11 & High Density Lipoprotein Cholesterol  \\
X12 & Low Density Lipoprotein Cholesterol  \\
X13 & Total Cholesterol  \\
X14 & Glucose  \\
X15 & Uric Acid  \\
X16 & Cystatin C  \\
X17 & C-Reactive Protein  \\
X18 & Glycated Hemoglobin  \\
X19 & Age  \\
X20 & Glomerular Filtration Rate  \\
X21 & Average Pulse Measure  \\
X22 & Body Mass Index  \\
X23 & Maximum Lung Function Peak Flow  \\
X24 & Cognitive Function Decrease (from the previous peak)  \\
\cline{1-2}
\hline
\end{tabular}
\label{tab15}
\end{center}
\end{table}
We select variables from three heterogeneous sources—physical exams, blood tests, and scale assessments—each exhibiting distinct distributions. Standard CI tests struggle with such cross-distribution dependencies. To address this, we adopt Petersen’s non-parametric CI test based on quantile regression and partial copulas~\cite{petersen_testing_2021}, which excels at detecting CI relationships among continuous variables with arbitrary distributions. However, its high computational cost necessitates dimensionality reduction before applying it to structural learning on even moderately sized graphs—precisely the setting where divide-and-conquer causal discovery is most valuable.
Conventional constraint-based divide-and-conquer methods first require costly CI tests to construct a Super-Structure or partition variables, especially in the absence of domain knowledge, rendering them impractical for such data. In contrast, our method constructs the Super-Structure \textbf{without any CI testing}, drastically reducing the total number of CI queries. This efficiency gain makes it feasible to deploy powerful but expensive non-parametric CI tests like Petersen’s in real-world, heterogeneous datasets.
\begin{figure}
    \centering
    \includegraphics[width=0.8\linewidth, height=0.4\linewidth]{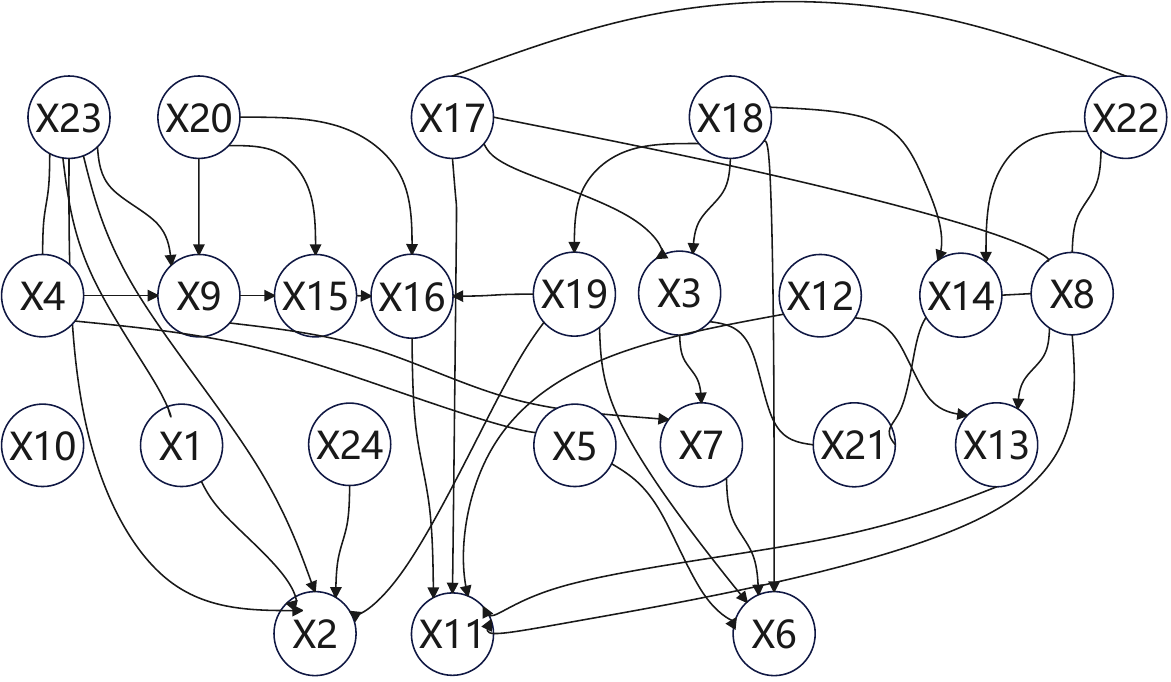}
    \caption{Results of algorithm execution}
    \label{fig:13}
\end{figure}
After applying Z-score normalization to all input data and handling missing values using Strobl’s method~\cite{strobl_fast_2018}, the final results are shown in Fig.~\ref{fig:13}.Algorithmic analysis results indicate that cognitive scores are influenced by factors such as depressive symptoms, prior decline in cognitive scores, age, and lung capacity. These findings align with existing consensus in cognitive science: cognitive function tends to decline with age in middle-aged and older adults~\cite{su_factors_2022}. Since cognitive scores essentially reflect an individual's cognitive ability, which is closely related to educational attainment, and educational attainment influences the trajectory of cognitive decline with age in middle-aged and older populations~\cite{li_associations_2020}. Depressive symptoms show a significant association with cognitive function and are an important risk factor~\cite{sun_correlation_2024}. Lung capacity can partially reflect the physical activity levels of middle-aged and older adults, and regular exercise has been proven to slow the progression of Alzheimer's disease~\cite{jia_effects_2019}. By comparing algorithmic learning outcomes with domain knowledge, it was found that the algorithm can, to some extent, reconstruct the causal relationships in the data.

\section{Conclusion}

We present a lightweight causal discovery framework that integrates weakly constrained Super-Structures with a divide-and-conquer strategy, enabling low-cost construction without domain knowledge—making it well-suited for large-scale, exploratory medical research. Our work provides a clear methodological contribution by formalizing the pipeline into three stages: Super-Structure construction, subgraph partitioning, and merging. Extensive experiments on synthetic, benchmark, and real-world datasets demonstrate its robustness and practical applicability. Although not yet state-of-the-art in accuracy, the framework offers significant computational advantages and opens new avenues for simplifying causal discovery and dimensionality reduction, encouraging a re-examination of the role of d-separation in divide-and-conquer approaches. Future work will focus on improving the efficiency and fidelity of weakly constrained Super-Structure construction and integrating advanced learning and merging strategies for broader applicability.


\bibliographystyle{IEEEtran}
\bibliography{ref}

\vspace{12pt}

\end{document}